%% file: root.tex
\documentclass[letterpaper, 10 pt, journal, twoside]{IEEEtran}

\input{packages.tex}

\begin{document}

\title{Aerial Gym Simulator: A Framework for Highly Parallelized Simulation of Aerial Robots}

% \author{Michael~Shell,~\IEEEmembership{Member,~IEEE,}
%         John~Doe,~\IEEEmembership{Fellow,~OSA,}
%         and~Jane~Doe,~\IEEEmembership{Life~Fellow,~IEEE}% <-this % stops a space
% \thanks{M. Shell was with the Department
% of Electrical and Computer Engineering, Georgia Institute of Technology, Atlanta,
% GA, 30332 USA e-mail: (see http://www.michaelshell.org/contact.html).}% <-this % stops a space
% \thanks{J. Doe and J. Doe are with Anonymous University.}% <-this % stops a space
% \thanks{Manuscript received April 19, 2005; revised August 26, 2015.}}
\author{Mihir Kulkarni, Welf Rehberg, and Kostas Alexis%
\thanks{Manuscript received: November, 02, 2024; Revised January, 23, 2025; Accepted February, 15, 2025.}%Use only for final RAL version
\thanks{This paper was recommended for publication by Editor  Giuseppe Loianno upon evaluation of the Associate Editor and Reviewers' comments.
This work was supported by the AFOSR Award No. FA8655-21-1-7033 and the Horizon Europe Grant Agreement No. 101119774. \textit{(Corresponding author: Mihir Kulkarni)}} %Use only for final RAL version
\thanks{All authors are with the Department of Engineering Cybernetics at the Norwegian University of Science and Technology, O.S. Bragstads Plass 2D, 7034, Trondheim, Norway ({e-mails: \{\tt\footnotesize mihir.kulkarni, welf.rehberg, konstantinos.alexis\}@ntnu.no}).}%
\thanks{Digital Object Identifier (DOI): see top of this page.}
}


\markboth{IEEE Robotics and Automation Letters. Preprint Version. Accepted February, 2025}
{Kulkarni \MakeLowercase{\textit{et al.}}: Aerial Gym Simulator: A Framework for Highly Parallelized Simulation of Aerial Robots} 

\maketitle
\input{definitions_and_abbreviations.tex}

\input{sections/abstract.tex}


\IEEEpeerreviewmaketitle

%%%%%%%%%%%%%%%%%%%%%%%%%%%%%%%%%%%%%%%%%%%%%%%%%%%%%%%%%%%%%%%%%%%%%%%%%%%%%%%%
\input{sections/introduction.tex}
%%%%%%%%%%%%%%%%%%%%%%%%%%%%%%%%%%%%%%%%%%%%%%%%%%%%%%%%%%%%%%%%%%%%%%%%%%%%%%%%
\input{sections/related_work.tex}
%%%%%%%%%%%%%%%%%%%%%%%%%%%%%%%%%%%%%%%%%%%%%%%%%%%%%%%%%%%%%%%%%%%%%%%%%%%%%%%%
\input{sections/features.tex}

%%%%%%%%%%%%%%%%%%%%%%%%%%%%%%%%%%%%%%%%%%%%%%%%%%%%%%%%%%%%%%%%%%%%%%%%%%%%%%%%
\input{sections/evaluation.tex}
%%%%%%%%%%%%%%%%%%%%%%%%%%%%%%%%%%%%%%%%%%%%%%%%%%%%%%%%%%%%%%%%%%%%%%%%%%%%%%%%
\input{sections/conclusions.tex}
%%%%%%%%%%%%%%%%%%%%%%%%%%%%%%%%%%%%%%%%%%%%%%%%%%%%%%%%%%%%%%%%%%%%%%%%%%%%%%%%

\bibliographystyle{IEEEtran}

\bibliography{root.bib}
\end{document}

%% file: packages.tex
\usepackage{graphics} % for pdf, bitmapped graphics files
\usepackage{epsfig} % for postscript graphics files
\usepackage{times} % assumes new font selection scheme installed
\usepackage{amsmath} % assumes amsmath package installed
\usepackage{amssymb}  % assumes amsmath package installed
\usepackage[T1]{fontenc}
\usepackage{graphicx}
\usepackage{color}
\usepackage{cite}
\usepackage{flushend}
\usepackage{url}
\usepackage[breaklinks,hidelinks]{hyperref}
\usepackage[capitalise]{cleveref}
\usepackage{booktabs}
\usepackage{makecell}
\usepackage{multirow}
\usepackage[nolist,nohyperlinks]{acronym}
\usepackage{bookmark}
\usepackage{bm}
\usepackage{tabularx}
\usepackage{pifont}
\usepackage{subcaption}
\usepackage{float}
\usepackage{derivative}

\usepackage{cleveref}
\usepackage{soul}

%% file: definitions_and_abbreviations.tex
\acrodef{uav}[UAV]{Uncrewed Aerial Vehicle}
\acrodef{drl}[DRL]{Deep Reinforcement Learning}
\acrodef{rl}[RL]{Reinforcement Learning}
\acrodef{ros}[ROS]{Robot Operating System}
\acrodef{gpu}[GPU]{Graphics Processing Unit}
\acrodef{cpu}[CPU]{Central Processing Unit}
\acrodef{gtd}[GTD]{Global Tensor Dictionary}

\newcommand{\ags}{Aerial Gym Simulator}
\newcommand{\isaacgym}{Isaac Gym}

\newcommand{\cmark}{\textcolor{green}{\text{\ding{51}}}}
\newcommand{\xmark}{\textcolor{red}{\text{\ding{55}}}}

\newcommand{\addition}[1]{#1}
\newcommand{\deletion}[1]{}

%% file: sections/abstract.tex
\begin{abstract}
\addition{This paper contributes the Aerial Gym Simulator, a highly parallelized, modular framework for simulation and rendering of arbitrary multirotor platforms based on NVIDIA Isaac Gym. Aerial Gym supports the simulation of \mbox{under-,} fully- and over-actuated multirotors offering parallelized geometric controllers, alongside a custom GPU-accelerated rendering framework for ray-casting capable of capturing depth, segmentation and vertex-level annotations from the environment.
Multiple examples for key tasks, such as depth-based navigation through reinforcement learning are provided. The comprehensive set of tools developed within the framework makes it a powerful resource for research on learning for control, planning, and navigation using state information as well as exteroceptive sensor observations. Extensive simulation studies are conducted and successful sim2real transfer of trained policies is demonstrated. The Aerial Gym Simulator is open-sourced at:\\
{\url{https://github.com/ntnu-arl/aerial_gym_simulator}}}.
\end{abstract}

%% file: sections/introduction.tex
\section{INTRODUCTION}

\IEEEPARstart{W}{ith} increasing deployment in a vast range of applications, including inspection, delivery, and search-and-rescue, aerial robots have gained immense popularity. Multirotor systems of varying scales have taken diverse roles and forms ranging from large vehicles with significant payload-carrying capacity to racing micro drones and reconfigurable robots capable of changing their shape actively or passively for traversal~\cite{zhao2018dragon,kulkarni2020reconfigurable,folding_drone,bucki2019design} or manipulation~\cite{zhao2017whole,huan@reconfigurable}. Critically, each unique robot configuration requires addressing embodiment- and task-specific challenges in terms of control, sensing capabilities, perception, and planning. With changes in the number of propellers, structural materials, overall platform size, payloads, the onboard sensor suite, as well as the environment within which a system is expected to operate, autonomy design and optimization need to exploit high-end simulation toward a safer and faster path to resilient deployment. This is even more critical when a) unconventional airframes are employed, such as novel multirotor configurations, reconfigurable multi-linked systems, or soft drones, as well as when b) learned exteroception-driven autonomous (e.g., depth cameras-based) navigation is concerned.

\begin{figure}[t!]
    \centering
    \includegraphics[width=0.98\columnwidth]{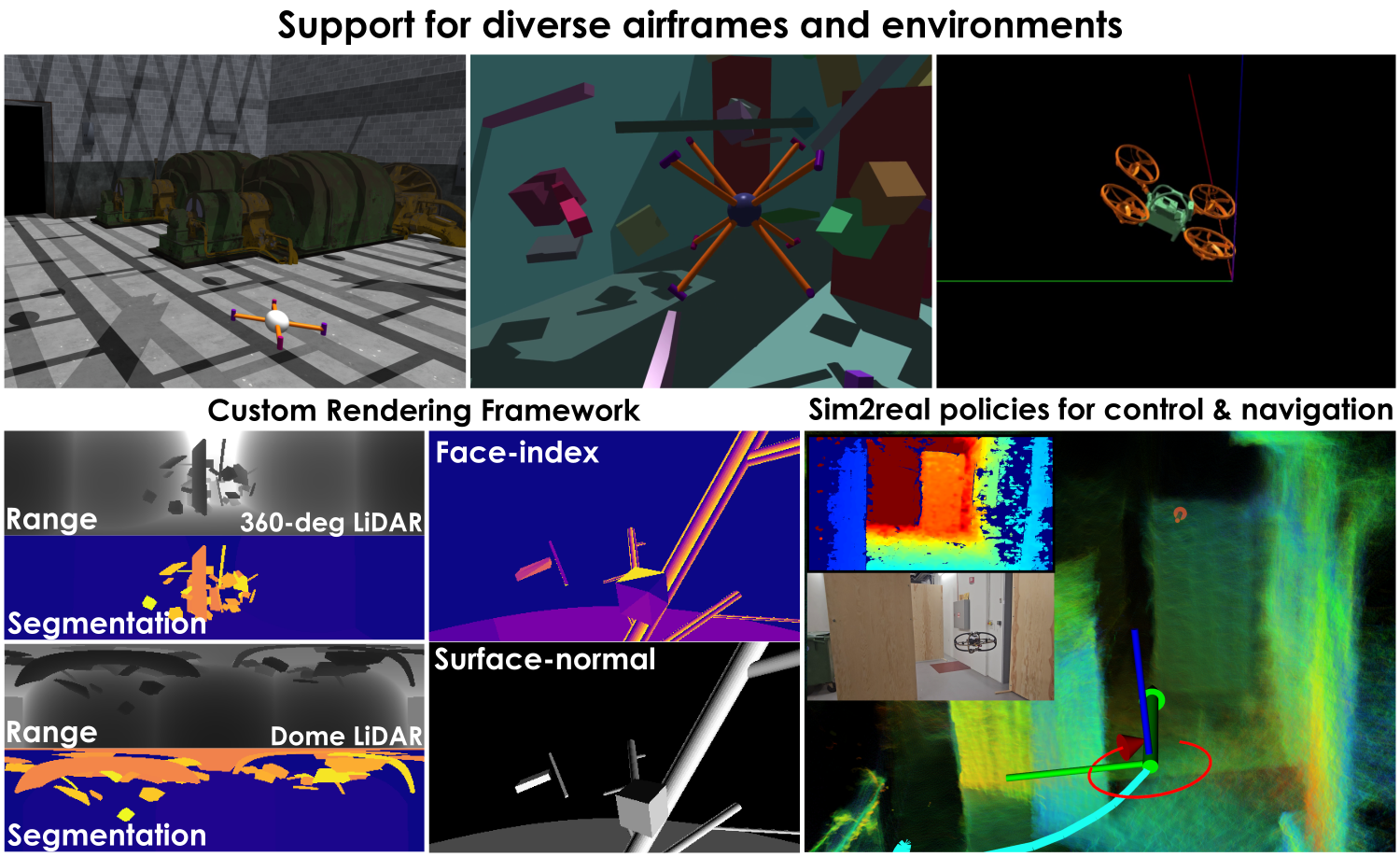}
    \vspace{-1.25ex}
    \caption{Salient features of the~\ags.}
    \label{fig:intro-figure}
    \vspace{-3ex}
\end{figure}

Accordingly, the field of aerial robotics shall greatly benefit from simulation tools that offer accurate physics models, parallelized rendering capabilities, out-of-the-box support for diverse airframes, and integration with established robot learning tools alongside ready-to-use control implementations.
To address these issues, we contribute the~\ags~which builds upon NVIDIA~\isaacgym~\cite{makoviychuk2021isaac}~and offers a feature-rich, modular, integrated, highly parallelized simulation and rendering framework that further incorporates explicit functionality for learning control and exteroceptive navigation policies as shown in~\cref{fig:intro-figure}. In detail, our contributions include:
\addition{
\vspace{-0.5ex}
\begin{itemize}
    \item \textbf{A highly-parallelized simulation} framework for simulation of under-, fully-, and over-actuated multirotor platforms.
    \item \textbf{A comprehensive control suite} and interfaces for arbitrary airframes across a wide range of abstraction levels ranging from position setpoints to RPM control with simulated motor dynamics.
    \item \textbf{A GPU-based rendering framework} to create, maintain, and update parallel rendering environments that can be randomized separately and modified at runtime. Standalone kernels for parallelized ray-casting are developed to simulate custom sensor models.
    \item \textbf{Integration with~\ac{drl} frameworks} exploiting ready-to-use-controllers, accelerated rendering and their inter-compatibility with various airframes. Established frameworks~\cite{rlgames2021,petrenko2020sample,huang2022cleanrl} are integrated for seamless policy training. Extensive experimental studies are conducted to demonstrate the sim2real transferability of policies trained for state-based control and vision-based navigation tasks.
\end{itemize}
}

In the remainder of this paper,~\cref{sec:related_work} presents related work, followed by the simulator description in~\cref{sec:features}. \cref{sec:benchmarking} provides benchmarks for physics and rendering throughput. Real-world and simulation experiments are detailed in~\cref{sec:experimental_evaluation} followed by conclusions in~\cref{sec:conclusions}.

%% file: sections/related_work.tex
\section{RELATED WORK} \label{sec:related_work}

High-fidelity simulators allow researchers to reliably test their algorithms in a safe, scalable, cost- and time-efficient manner. There has been immense progress in the field of aerial robot simulation with a wide number of simulators being used~\cite{dimmig2024survey}. Several simulators respond to individual needs such as that of accurate dynamics (RotorS~\cite{furrer2016rotors} and Hector Quadrotor~\cite{hectorquadrotor2012meyer}) by developing on top of universal simulators like Gazebo~\cite{koenig2004gazebosim}. Other simulators such as FlightGoggles~\cite{guerra2019flightgoggles} and Flightmare~\cite{song2020flightmare} focus on photorealistic rendering and high-fidelity graphics using Unity~\cite{haas2014history}, without simulating contact or collisions. AirSim~\cite{shah2017airsim}, built over the Unreal Engine~\cite{unrealengine} simulates robot dynamics and collisions while offering high-fidelity rendering but typically requires large computational resources. Gym-pybullet-drones~\cite{panerati2021gym_pybullet_drones} developed over the PyBullet~\cite{coumans2016pybullet} physics engine, offers CPU-based simulation and rendering.

With the advent of \ac{gpu}-accelerated physics and rendering engines, simulators and frameworks such as Isaac Gym~\cite{makoviychuk2021isaac}, Orbit~\cite{mittal2023orbit} and MuJoCo~\cite{todorov2012mujoco} are widely used to train policies for multi-linked legged systems and articulated robotic arms, often demonstrating impressive locomotion and manipulation capabilities~\cite{rudin2022learningtowalk,hoeller2024parkour,huan2024leggedtail,factory2022narang} with proprioceptive and exteroceptive inputs. Similar work related to control, planning, navigation and manipulation with multi-linked flying and soft systems~\mbox{\cite{huan@reconfigurable,kulkarni2020reconfigurable,zhao2017whole,paolo2024morphy,zhao2018dragon}} lacks available integration with simulation tools for ease of reliable replication and reproduction. \addition{Notably, OmniDrones~\cite{omnidrones} based on NVIDIA Isaac Sim provides support for the simulation of reconfigurable multirotor platforms based on motor commands but lacks support for control-allocation for non-planar airframes. While the native rendering functionality offered with Isaac Sim is slow, Warp based functionality allows for faster ray-casting against a single mesh that is shared across parallel environments and cannot be modified.}

The proposed open-sourced simulation framework aims to bridge the gap by providing interfaces for highly parallelized simulation of aerial systems, closely integrated with a custom \ac{gpu}-accelerated rendering framework. Providing interfaces that imitate real-world robot interfaces at various levels of control abstraction enables rapid development for policies using both proprioceptive and exteroceptive sensing modalities for control, navigation, and planning for arbitrary multirotor configurations. Moreover, a set of provided examples for training policies for depth-based autonomous navigation and control for arbitrary multirotor platforms allow users to jump-start their research in learning for multirotor platforms.

%% file: sections/features.tex
\section{FEATURES}\label{sec:features}

Exploiting the physics engine and interfaces provided by NVIDIA~\isaacgym~and developing on the advancements made in an earlier pre-release~\cite{kulkarni2023aerial}, this simulator represents a mature solution with a host of new features. \addition{A rich set of tools is provided that enables massively parallel simulation for aerial platforms, an integrated GPU-accelerated rendering framework to simulate custom sensor models and new sensing modalities with support for multirotor configurations with arbitrary motor numbers and locations.} The simulator is redesigned, allowing each module to access the latest states and sensor data, leading to increased rendering throughput and overall simulation speed. A set of parallelized controllers at various control abstraction levels imitating real-world control interfaces are provided, alongside modeling of motor dynamics to bridge the sim2real gap. \addition{Additionally, linear and quadratic terms can be configured to simulate aerodynamic drag. Finally, the combined features of the simulator enable researchers to jump-start training policies for control, planning, and navigation for multirotor platforms with exteroceptive sensor measurements such as depth maps.}

\input{sections/features/architecture.tex}

\input{sections/features/multirotor_embodiments.tex}

\input{sections/features/controllers.tex}

\input{sections/features/sensors.tex}

\input{sections/features/drl_navigation.tex}

%% file: sections/features/architecture.tex
\subsection{Architecture}

\begin{figure*}[!h]
    \centering
    \includegraphics[width=0.98\textwidth]{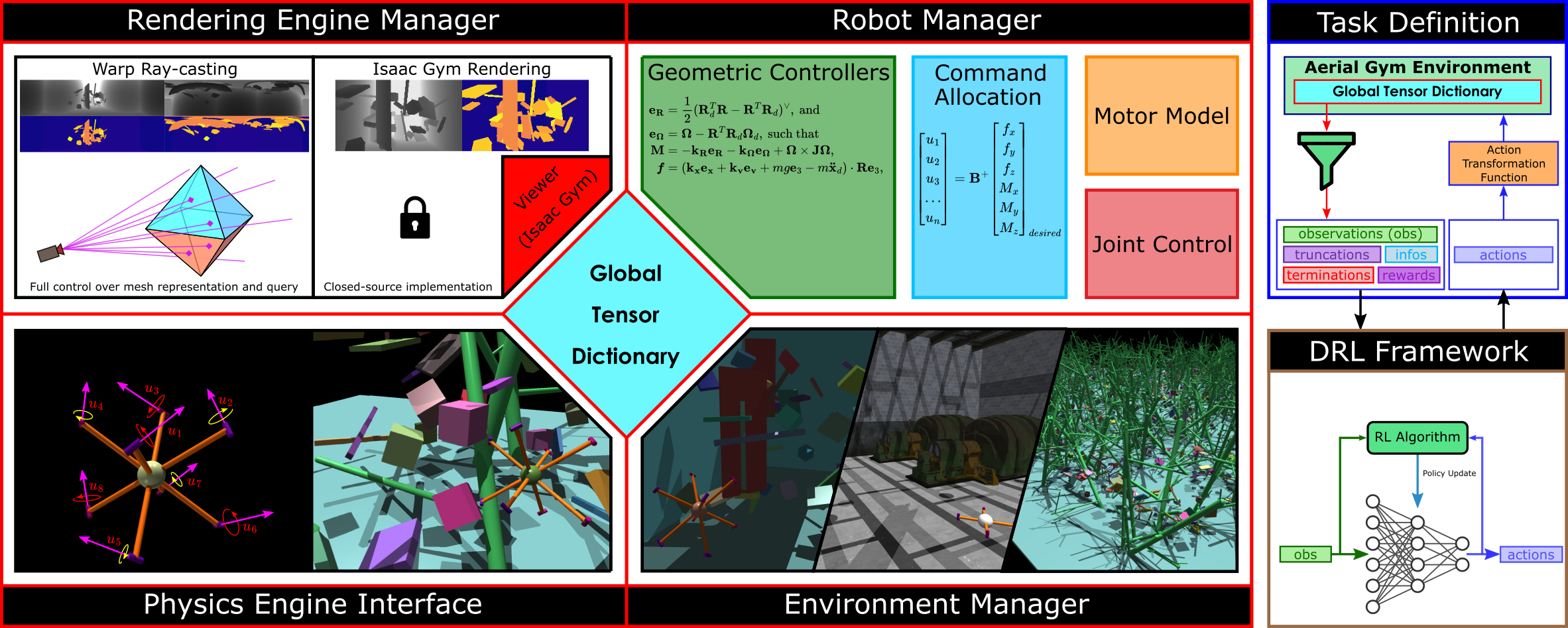}
    \caption{Core components for the~\ags~include the physics engine (\isaacgym), Warp and Isaac Gym-based rendering solutions, robot control suite and tools for environment generation and randomization.}
    \label{fig:simulator_architecture}
    \vspace{-2ex}
\end{figure*}

The simulator is designed to be modular with separate components for each functionality as shown in~\cref{fig:simulator_architecture}. The major components of the~\ags~are the managers for the rendering engine, robots and environments, interfaces for the physics engine and common~\ac{drl} frameworks along with the definition of learning tasks. All these components share a common memory bank dubbed as the~\ac{gtd} and perform in-place operations on the tensors in~\ac{gtd}. The managers for rendering engine, robots and environments perform the substeps related to their domain. Aerial Gym obtains states and joint information of all simulated entities from the~\isaacgym~physics engine and updates the~\ac{gtd} at each physics step. The geometric controllers use the robot states from the~\ac{gtd} and provide the forces and torques to be applied to the robot's actuators. \addition{Task definitions} are constructed conforming to the Gymnasium API~\cite{kwiatkowski2024gymnasium} and provide task-specific interpretations of the information stored in the~\ac{gtd}. These are then provided as observations to an interfaced~\ac{drl} framework for policy training.

%% file: sections/features/multirotor_embodiments.tex
\subsection{Multirotor Embodiments}\label{subsec:multirotor_embodiments}

\begin{figure}[!htbp]
    \vspace{-2ex}
    \centering
    \includegraphics[width=0.98\columnwidth]{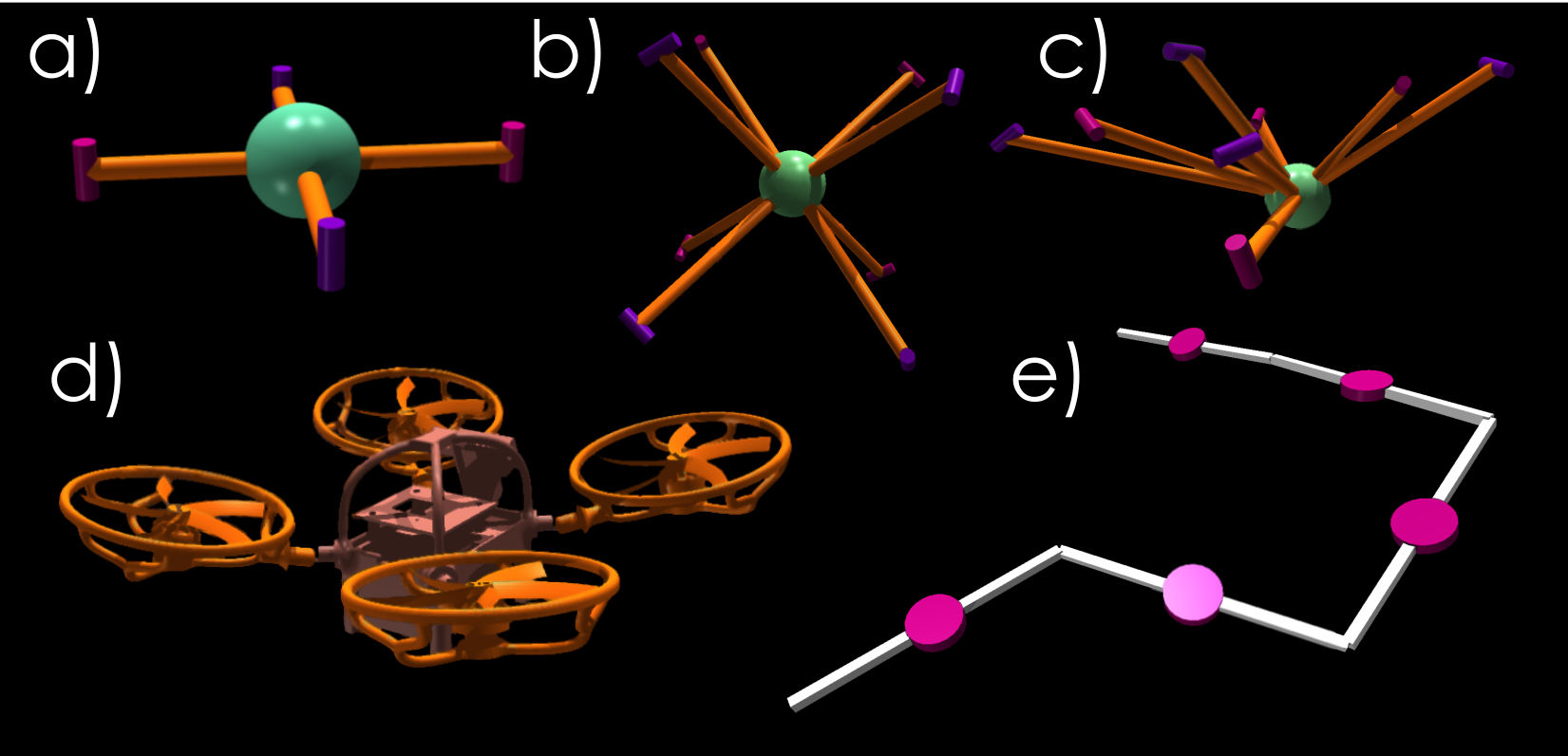}
    \caption{Various airframes provided with the~\ags. Rigid quad- (a) and octa-rotors (b and c), alongside a compliant robot \textit{Morphy} (d) based on~\cite{paolo2024morphy} and reconfigurable robot (e) inspired by~\cite{zhao2017whole}.}
    \label{fig:embodiments}
    \vspace{-2ex}
\end{figure}

With renewed interest in task-specific embodiments, including non-planar airframes~\cite{brescianini2016octarotor} and highly asymmetric multi-linked systems, that may further involve reconfigurability~\cite{zhao2017whole,zhao2018dragon,huan@reconfigurable} and softness~\cite{paolo2024morphy}, there is a need for simulators that support such platforms. Responding to this need, the~\ags~is developed to support the simulation of various airframe configurations out-of-the-box. A user-specified configuration file for each robot platform defines the number of motors, joint parameters and the selected sensors for an embodiment defined with a Universal Robot Description Format (URDF) file. The simulator is also able to handle complex meshes and convex decompositions of non-convex meshes to simulate collisions with the environment albeit at a higher computation cost.~\cref{fig:embodiments} shows example robot airframes provided with the simulator.

\subsubsection{Simulation for Arbitrary Rigid Configurations}\label{subsec:arbitrary_multirotor}

Arbitrary $n$-motor configurations can be simulated, \addition{with a robot configuration file specifying the number, relative poses, and directions of the motors. Robots are defined using URDF files as required by NVIDIA Isaac Gym.}

\subsubsection{Flexible and Reconfigurable Airframes}\label{subsec:flexible_joints}

Recent works investigate actively and passively morphing multirotors with flexible arms that change the robot morphology in flight~\mbox{\cite{folding_drone,zhao2018dragon,bucki2019design}}. The~\ags~supports simulation of such airframes with both active and passive rotational joints. Each joint can be interfaced with a PD-controller provided by~\isaacgym, or with a custom implementation matching the actuator or the compliant-response properties of real-world robots. \addition{Open-source examples include an accurate model of compliant joints of the Morphy platform~\cite{paolo2024morphy}, alongside a model for a reconfigurable multi-linked flying platform inspired by the work in~\cite{zhao2017whole} to accelerate relevant research efforts.
}

%% file: sections/features/controllers.tex
\subsection{Controllers and Interfaces}\label{subsec:controllers}

The simulator is equipped with various controllers across levels of abstraction based on the taxonomy in~\cite{eschmann2024learning}, intended to accelerate research efforts by exploiting commonly-used control interfaces with learning algorithms. Adapted geometric controllers based on~\cite{lee2011control} are provided, with sub-optimal performance on non-planar configurations serving as a starting point for future research. The provided controllers employ PyTorch's JIT-compiled modules for faster execution on the~\ac{gpu}. Robots with any chosen level of control abstraction can be closely integrated with a~\ac{drl} framework to train policies for control and navigation. The following controllers, interfaces are currently provided:

\subsubsection{Attitude-Thrust and Angular Rate-Thrust Controllers}

The errors in desired orientation $\mathbf{e_R}$ and body rates $\mathbf{e}_{\boldsymbol{\Omega}}$, alongside the resulting desired body-torque $\mathbf{M}$ are computed as per~\cite{furrer2016rotors} which is inspired by~\cite{lee2011control}:
\vspace{-1ex}
\begin{align}
    \mathbf{e_R} &= \frac{1}{2}  (\mathbf{R}_d^T  \mathbf{R} - \mathbf{R}^T  \mathbf{R}_d)^{\vee}, \\
    \mathbf{e}_{\boldsymbol{\Omega}} &= \boldsymbol{\Omega} - \mathbf{R}^T  \mathbf{R}_d  \boldsymbol{\Omega}_d,~\textrm{and} \\
    \mathbf{M} &= -\mathbf{k_R} \mathbf{e_R} - \mathbf{k}_{\boldsymbol{\Omega}} \mathbf{e}_{\boldsymbol{\Omega}} + \boldsymbol{\Omega} \times \mathbf{J} \boldsymbol{\Omega}\label{eq:moment_eqn},
\end{align}
where $\mathbf{M}$ is the desired body-torque expressed in the body-fixed frame $\mathcal{B}$, $\mathbf{R}$ and $\mathbf{R}_d$ denote the current and desired orientation, ${\boldsymbol{\Omega}}$ and ${\boldsymbol{\Omega}}_d$ denote current and desired angular rates of the robot, all expressed in $\mathcal{B}$ and $\boldsymbol{k_R,k_{\boldsymbol{\Omega}}}$ being adequate weights. The robot moment of inertia is denoted as $\mathbf{J}$ and the vee-map is denoted by $\cdot^{\vee}$. For attitude-thrust control, the desired angular velocity is set to $\boldsymbol{\Omega}_d = 0$, and for angular rate-thrust control $\mathbf{R}_d = \mathbf{R}$. The total thrust command $\mathbf{f}$ expressed in $\mathcal{B}$ is directly provided to the control allocation scheme to obtain motor commands.

\subsubsection{Position, Velocity and Acceleration Controllers}

The desired body-torque $\mathbf{M}$ is calculated as in (\ref{eq:moment_eqn}) and the thrust command $\mathbf{f}$ is calculated as:

\vspace{-3ex}
\small
\begin{align}
    \mathbf{f} &= (\mathbf{k_x} \mathbf{e_x} + \mathbf{k_v} \mathbf{e_v} + m g \mathbf{e}_3 - m \mathbf{\ddot{x}}_d) \cdot \mathbf{R}\mathbf{e}_3,
\end{align}
\normalsize
where $\mathbf{e_x}$ and $\mathbf{e_v}$ denote the position and velocity errors in the inertial frame $\mathcal{W}$, $\mathbf{k_x}~\textrm{and}~\mathbf{k_v}$ denote the respective weights, $g$ is the magnitude of acceleration due to gravity, $m$ is the robot mass, $\ddot{\mathbf{x}}_d$ is the desired robot acceleration in $\mathcal{W}$, while $\mathbf{e}_3$ is a unit vector in the \textit{z}-direction. The matrix denoting desired orientation in this case is calculated as $\mathbf{R}_d = [ \mathbf{b}_{2_c} \times \mathbf{b}_{3_c}; \mathbf{b}_{2_c}; \mathbf{b}_{3_c}]$, where

\vspace{-2ex}
\small
\begin{align}
    \mathbf{b}_{3_c} = - \frac{-\mathbf{k_x} \mathbf{e_x} - \mathbf{k_v} \mathbf{e_v} - m g \mathbf{e_3} + m \mathbf{\ddot{x}}_d}{ || -\mathbf{k_x} \mathbf{e_x} - \mathbf{k_v} \mathbf{e_v} - m g \mathbf{e_3} + m \mathbf{\ddot{x}}_d ||_2},
\end{align}
\normalsize
and $\addition{\mathbf{\chi} = [\cos{\Phi_d}, \sin{\Phi_d}, 0]^T}$, such that \addition{$\mathbf{b}_{2_c} = (\mathbf{b}_{3_c} \times \mathbf{\chi})/||\mathbf{b}_{3_c} \times \mathbf{\chi}||_2$}, \addition{where $\Phi_d$ denotes the desired yaw} as per~\cite{furrer2016rotors}. The $\boldsymbol{\ell}^2$-norm is denoted by $||\cdot||_2$. For the case of velocity control $\mathbf{e_x} = \mathbf{0}$ and $\mathbf{\ddot{x}}_d = \mathbf{0}$, for acceleration control $\mathbf{e_x} = \mathbf{0}$ and $\mathbf{e_v} = \mathbf{0}$. Similarly, for position control $\mathbf{e_v} = \mathbf{0}$ and $\mathbf{\ddot{x}}_d = \mathbf{0}$.

\subsubsection{Body-wrench Controller}

The desired or externally-commanded body-wrench is represented as $\mathbf{W} = [{f}_x, {f}_y, {f}_z, {M}_x, {M}_y, {M}_z]^T$. For an $n$-rotor system, the thrust command provided to the actuators is $\mathbf{U} = [u_1, u_2, \dots, u_n]^T$. The embodiment-specific control-effectiveness matrix $\mathbf{B}$~\cite{johansen2013control} relates $\mathbf{U}$ and $\mathbf{W}$ as:

\vspace{-3ex}
\begin{align}
    \mathbf{W} &= \mathbf{B}\mathbf{U}, \textrm{and} \\
    \mathbf{U}^\textrm{ref} &= \mathbf{B}^+~{\mathbf{W}^\textrm{ref}}\label{eq:allocations}.
    \vspace{-4ex}
\end{align}
The Moore-Penrose inverse (pseudoinverse), represented as $\mathbf{B}^+$ is used to obtain an unconstrained least-squares solution for the desired or commanded motor forces $\mathbf{U}^\textrm{ref}$, given a desired wrench $\mathbf{W}^\textrm{ref}$. This allocation scheme is suboptimal for systems with input constraints but is a common approximation that can be replaced with custom strategies.

\subsubsection{RPM and Thrust Setpoint Interfaces}

The Aerial Gym Simulator provides functionalities to choose between the interface to command the motors via either thrust or RPM setpoints. The desired RPM at time $t$ for motor $k$, $r^{\textrm{ref}}_{k,t}$ is obtained by solving the equation $u^{ \textrm{ref}}_{k,t} = c_{f,k} ~ {r_{k,t}^\textrm{ref}}^2$, where $u^{ \textrm{ref}}_{k,t}$ is the reference thrust setpoint and $c_{f,k}$ is the thrust constant for motor $k$. \addition{The continuous-time response of the motor is modeled as:}

\vspace{-3ex}
\small
% \color{blue}
\begin{align}
\dot{r}_k &= \frac{1}{\tau_{k}}(r^{\textrm{ref}}_{k} - r_{k}), ~\textrm{where} \\
\tau_{k} &= \begin{cases}
    \tau^{\textrm{inc}}_{k}~~\textrm{if }r^{\textrm{ref}}_{k} \geq r_{k}\label{eq:motor_dynamics}\\
    \tau^{\textrm{dec}}_{k}~~\textrm{if }r^{\textrm{ref}}_{k} < r_{k}
    \end{cases}
\end{align}
\color{black}
\normalsize
\addition{The motor-time constant is $\tau_{k}$ and takes the value $\tau^{\textrm{dec}}_{k}$ in case of slowing down or $\tau^{\textrm{inc}}_{k}$ otherwise. The response can be simulated using both the RK4 and Euler methods for numerical integration.} The applied force is $u_{k, t}=c_{f,k} ~ r_{k, t}^2$ and the applied torque is equal to $-d_k ~ c_{\tau,k} ~ u_{k, t}$, where $ c_{\tau,k}$ is the torque coefficient, and $d_k$ is the direction of rotation.

%% file: sections/features/sensors.tex
\subsection{Sensors}\label{sec:sensors}

\begin{figure*}[ht!]
    \centering
    \begin{subfigure}[t]{0.60\textwidth}
        \centering
        \includegraphics[height=2.45in]{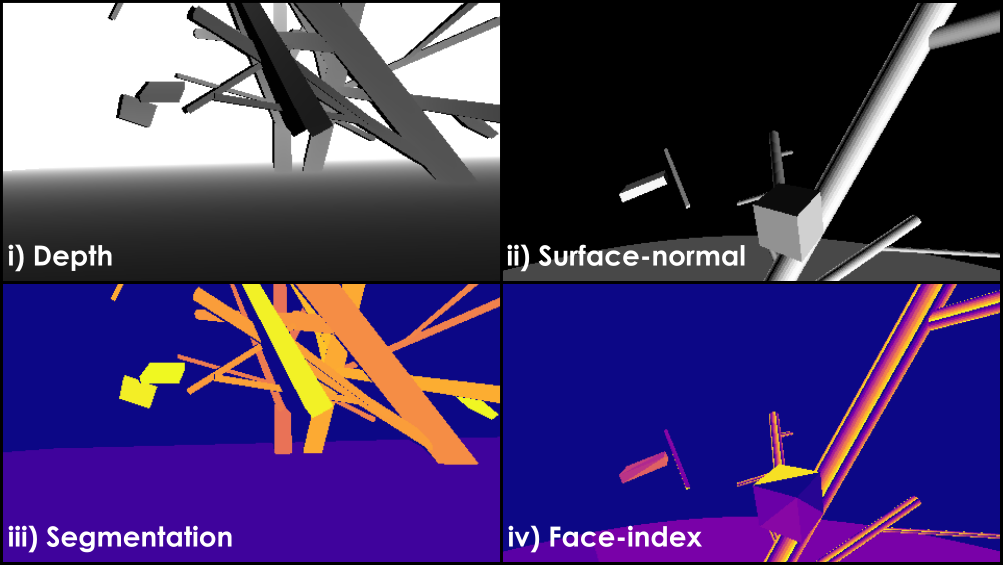}
        \caption{Camera Sensor (with resolution $270 \times 480$)}
        \label{subfig:camera_sensor}
    \end{subfigure}%
    ~~~~
    \begin{subfigure}[t]{0.40\textwidth}
        \centering
        \includegraphics[height=2.45in]{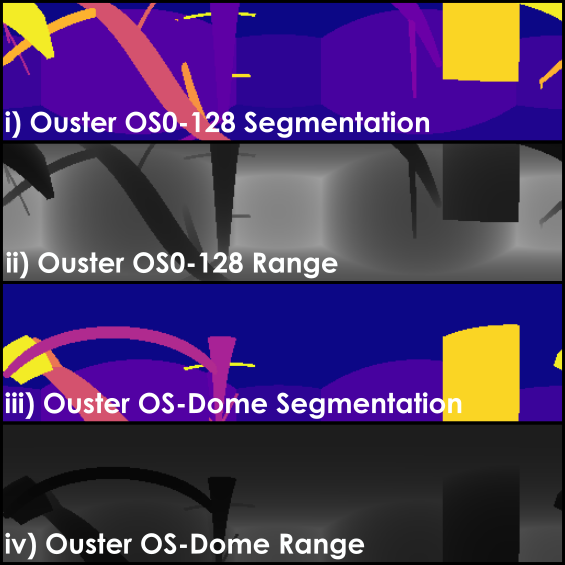}
        \caption{LiDAR Sensors ($512$ points, $128$ channels)}
        \label{subfig:lidar_sensor}
    \end{subfigure}
    \caption{Sensor measurements using the proposed rendering framework. \cref{subfig:camera_sensor} shows depth, segmentation, surface-normal and face-index images captured by a simulated camera sensor. \cref{subfig:lidar_sensor} shows the data captured using a standard $3$-D LiDAR and a hemispherical dome LiDAR sensor.}
    \vspace{-4ex}
\end{figure*}

In the following section, the simulated sensors provided by the~\ags~are described.

\subsubsection{Exteroceptive Sensors}\label{subsec:exteroceptive_sensors}

\addition{NVIDIA Isaac Gym comes packaged with a rendering solution that has several limitations. Specifically, the users are restricted to only camera sensors using a pinhole projection model, while providing interfaces for RGB, depth, segmentation and optical flow information.} The addition of \addition{data types} \addition{and different projection} models for sensors such as Time-of-Flight (ToF) cameras and $3$-D LiDARs is not possible. Moreover, the loop-based access scheme for each sensor instance provided by~\isaacgym~for pose randomization leads to slower simulation speeds. \addition{To overcome these limitations, we implement a custom rendering framework based on NVIDIA Warp~\cite{warp2022} that complements the default rendering engine.} For an environment $\boldsymbol{\mathcal{E}}_{i}$, a base triangle-mesh representation containing $n_i$ sub-meshes is created as $\boldsymbol{\mathcal{M}}^{\textrm{base}}_i =\{\mathbf{M}_{j}\}_{j \in [1, n_i]}$, where each sub-mesh $\mathbf{M}_j$ corresponds to an individual object in simulation. Simultaneously, transformations for each sub-mesh at time $t$ are maintained as $\boldsymbol{\mathcal{T}}_{i,t} = \{\mathbf{T}_{j, t}\}_{j \in \{1, \cdots, n_i\}}$. The vertices of the mesh are transformed to match the obstacles in the simulator at time $t$ as $\boldsymbol{\mathcal{M}}_{i,t} = \boldsymbol{\mathcal{T}}^{T}_{i,t} \circ \boldsymbol{\mathcal{M}}^{\textrm{base}}_{i}$, where the $\circ$ operator defines the transformation of each vertex of a sub-mesh $\mathbf{M}_j$ with its corresponding transformation matrix $\mathbf{T}_{j, t}$ for environment $\boldsymbol{\mathcal{E}}_i$. Subsequently, a bounding volume hierarchy is calculated for $\boldsymbol{\mathcal{M}}_{i,t}$ for efficient ray-casting. \addition{While computationally expensive, performing this exclusively for randomization of static environments allows the~\ags~to effectively create, maintain and update multiple independent environments.}

\addition{Standalone kernels for ray-casting are developed and integrated with the simulator. To obtain measurements, individual rays are cast outwards per-pixel to evaluate intersection with $\boldsymbol{\mathcal{M}}_{i,t}$. The distance between the sensor and the point-of-intersection is reported as range for ToF sensors and LiDARs, while the distance of this point from the image plane is reported as depth. Shadows observed in stereo-camera systems are simulated by projecting rays back from the points-of-intersection towards the second sensor and marking pixels corresponding to the rays that intersect with the environment as invalid~\cite{scharstein2002taxonomy}. Additionally, it is also possible to embed vertex-level annotations that can be queried to obtain user-defined information from the environment. For example, vertex (or mesh-face) indices alongside barycentric coordinates of intersecting rays can be used to obtain user-specified information and infer gradients along the mesh surface as needed. Our realization of the GPU-accelerated ray-casting offers an order of magnitude speedup compared to the default rendering engine of~\isaacgym~with support for newer data types such as point clouds and surface normals (\cref{subfig:camera_sensor})}. User-defined sensors with customizable projection models (e.g., Dome LiDAR~\cref{subfig:lidar_sensor}) can be added. Moreover, the custom tensor-based implementation allows for consolidation of image tensors and faster randomization of sensor poses compared to~\isaacgym's native loop-based API camera interface. For ease of use, we provide idealized configurations of: Ouster OS-0, OS-1, OS-2, OS-Dome, Intel RealSense D455, Luxonis Oak-D (and Pro W) and ST VL53L5CX ToF sensors.

\subsubsection{IMU Sensor}

An IMU sensor model is implemented in the~\ags~to obtain proprioceptive measurements simulating real-world sensors. The simulated IMU utilizes the true forces measured by the simulator alongside a Gaussian noise and a discrete-time bias random-walk implementation as described in~\cite{kalibr@2016}. The measured acceleration $\mathbf{a}_{\textrm{meas}}$ and angular velocity $\boldsymbol{\Omega}_{\textrm{meas}}$ about the sensor are calculated as:

% alignat environment allows setting two alignment anchor points
\vspace{-3ex}
\small
\begin{alignat}{2}
    \mathbf{a}_{\textrm{meas}, t} &= \mathbf{a}_{\textrm{true}, t} + \mathbf{b}_{\mathbf{a},t} + \mathbf{n_a};\quad &&\mathbf{n_a} \sim \mathcal{N}(\mathbf{0},\boldsymbol{\sigma}_\mathbf{a}), \\
    \boldsymbol{\Omega}_{\textrm{meas}, t} &= \boldsymbol{\Omega_}{\textrm{true}, t} + \mathbf{b}_{\boldsymbol{\Omega}, t} + \mathbf{n}_{\boldsymbol{\Omega}};\quad &&\mathbf{n}_{\boldsymbol{\Omega}} \sim \mathcal{N}(\mathbf{0},\boldsymbol{\sigma}_{\boldsymbol{\Omega}}),  \\
    \mathbf{b}_{\mathbf{a},t} &= \mathbf{b}_{\mathbf{a},t-1} + \mathbf{z}_{\mathbf{b}_{\mathbf{a}}};\quad &&\mathbf{z}_{\mathbf{b}_{\mathbf{a}}} \sim \mathcal{N}(\mathbf{0},\boldsymbol{\sigma}_{\mathbf{b_a}}), \\
    \mathbf{b}_{\boldsymbol{\Omega},t} &= \mathbf{b}_{\boldsymbol{\Omega},t-1} + \mathbf{z}_{\mathbf{b}_{\boldsymbol{\Omega}}};\quad &&\mathbf{z}_{\mathbf{b}_{\boldsymbol{\Omega}}} \sim \mathcal{N}(\mathbf{0},\boldsymbol{\sigma}_{\mathbf{b}_{\boldsymbol{\Omega}}}),
\end{alignat}
\normalsize
where $\mathbf{a}_{\textrm{true}, t} = \addition{\mathbf{R}_t^T} ((\mathbf{F}_{\textrm{true}, t} / m) + g\mathbf{e_3})$, and $ \mathbf{F}_{\textrm{true}, t}$ is the true net force expressed in $\mathcal{W}$ and $\boldsymbol{\Omega}_{\textrm{true}, t}$ is the angular rate experienced by the simulated robot at time $t$ expressed in $\mathcal{B}$, $\mathbf{b}_{\mathbf{a},t}$ and $\mathbf{b}_{\boldsymbol{\Omega},t}$ are the biases and $\mathbf{n}_{\mathbf{a}, t}$ and $\mathbf{n}_{\boldsymbol{\Omega}, t}$ are the noises for accelerometer and gyroscope sampled using a Gaussian distribution $\mathcal{N}$ with standard deviations $\boldsymbol{\sigma}_\mathbf{a}$ and $\boldsymbol{\sigma}_{\boldsymbol{\Omega}}$ respectively. $\mathbf{z}_{\mathbf{b}_{\mathbf{a}}}$ and $\mathbf{z}_{\mathbf{b}_{\boldsymbol{\Omega}}}$ are the bias-drift terms while $\boldsymbol{\sigma}_{\mathbf{b_a}}$ and  $\boldsymbol{\sigma}_{\mathbf{b}_{\boldsymbol{\Omega}}}$ are the standard deviations of the discrete-time random-walk model for the accelerometer and gyroscope respectively. The simulated IMU orientation can be randomized with the measurements appropriately transformed. Noise density and bias random-walk configurations are measured and provided for the VectorNav VN-100 and Bosch BMI085 IMUs.

%% file: sections/features/drl_navigation.tex
\subsection{Learning-based Control and Navigation}\label{subsec:drl_nav}

Deep neural networks have shown great promise in tackling the challenge of state-based control~\cite{eschmann2024learning,rudin2022learningtowalk,kaufmann2023champion}. However, effective training with high-dimensional exteroceptive sensor data remains a pertinent challenge. In view of accelerating this effort, we integrate established learning frameworks such as RL Games~\cite{rlgames2021}, Sample Factory~\cite{petrenko2020sample} and Clean RL~\cite{huang2022cleanrl} with Aerial Gym. Interfaces are provided to train policies for 
multirotor control and exteroceptive sensor-based navigation. Environments conforming to the Gymnasium standard~\cite{kwiatkowski2024gymnasium} are provided for control and navigation tasks for aerial platforms at varying levels of control abstraction. \addition{To deliver ready-to-use tools and examples we a) package environments for~\ac{drl}-based control of arbitrary multirotor platforms for position-setpoint tracking with deep neural networks interfaced to command robots at the chosen control-abstraction level, b) exploit the integrated rendering framework offering exteroceptive sensor data to train~\ac{drl}~policies for map-free navigation and c) demonstrate through experimental evaluations robust sim2real transfer of policies trained using the simulator.} Results from experimental validations are shown in~\cref{sec:experimental_evaluation}.

%% file: sections/evaluation.tex
\section{BENCHMARKING}\label{sec:benchmarking}

\input{sections/evaluation/benchmarking.tex}

\section{Experimental Evaluation}\label{sec:experimental_evaluation}

\input{sections/evaluation/experiments.tex}

%% file: sections/evaluation/benchmarking.tex
\begin{table*}[h]
  \centering
  \caption{Simulation Throughput (samples per second (SPS) \& frames per second (FPS))  and Feature Comparison}\label{tab:speed_benchmarks}
  \begin{scriptsize}
  \begin{tabular}{| c | cccc | cccc | cccc |}
      \hline
        & \multicolumn{4}{c|}{\textbf{Physics (SPS)} $(\times 10^6)$} & \multicolumn{4}{c|}{\textbf{Rendering (FPS)}} & \multicolumn{4}{c|}{\textbf{Sensors}} \\ \hline
      \textbf{Instances / Features} & $\mathbf{2}^\mathbf{4}$ & $\mathbf{2}^\mathbf{8}$ & $\mathbf{2}^\mathbf{12}$ & $\mathbf{2}^\mathbf{16}$ & $\mathbf{2}^\mathbf{4}$ & $\mathbf{2}^\mathbf{6}$ & $\mathbf{2}^\mathbf{8}$ & $\mathbf{2}^\mathbf{10}$ & \textbf{RGB-D(Seg)} & \textbf{LiDAR} & \textbf{IMU} & \textbf{Face/Normal} \\ \hline
      gym-pybullet-drones~\cite{panerati2021gym_pybullet_drones} & $0.003$ & $0.004$ & $0.004$ & \xmark & $16.70$ & $5.13$ & \xmark & \xmark & \cmark & \xmark & \xmark & \xmark \\ \hline
      L2F~\cite{eschmann2024learning}$^{*}$ & $\mathbf{16.32}$ & $\mathbf{266.66}$ & $\mathbf{4311}$ & $\mathbf{14063}$ & \xmark & \xmark & \xmark & \xmark & \xmark & \xmark & \xmark & \xmark \\ \hline
      OmniDrones~\cite{omnidrones} (Native) & \multirow{2}{*}{$0.002$}  & \multirow{2}{*}{$0.038$} & \multirow{2}{*}{$0.450$} & \multirow{2}{*}{\xmark} & $219$ & $281$ & $307$ & \xmark & \cmark & \xmark & \xmark & \xmark/\cmark \\
      OmniDrones~\cite{omnidrones} (Warp) &  & & & & $400$ & $1037$ & $1770$ & \xmark & \xmark & \cmark & \xmark & \xmark \\ \hline
      \textbf{\ags} & $0.007$ & $0.130$ & $1.43$ & $4.43$ & $\mathbf{1404}$ & $\mathbf{2370}$ & $\mathbf{2714}$ & $\mathbf{3921}$ & \cmark & \cmark & \cmark & \cmark \\ \hline
    \end{tabular}
  \end{scriptsize}
  % \vspace{-2ex}
\end{table*}

Benchmarking studies are performed against relevant open-source simulators that support the simulation of multiple multirotor platforms with either CPU or GPU-based simulation and rendering. We benchmark and compare the physics simulation speeds and rendering throughput of the \ags~against gym-pybullet-drones~\cite{panerati2021gym_pybullet_drones}, OmniDrones~\cite{omnidrones}, and the simulator in Learning to Fly in Seconds (L2F)~\cite{eschmann2024learning}. The comparison is performed by commanding constant RPM setpoints to quadrotors in obstacle-free environments for all simulators. Similarly, rendering performance is measured for the simulators in similar environments consisting of $20$ cube obstacles in front of the robots. A single camera with a resolution of $270 \times 480$ pixels is simulated per robot capturing both depth and segmentation images. A workstation with an AMD Ryzen Threadripper Pro 3975WX CPU and NVIDIA RTX 3090 GPU is used for comparisons. The results for physics throughput in samples per second (SPS) and rendering throughput in frames per second (FPS) are tabulated in~\cref{tab:speed_benchmarks} alongside the provided rendering capabilities. An unsupported feature or absence of data due to simulator crash is indicated by~\xmark, while ${*}$ indicates values reported by the authors of L2F (using a laptop GPU).

\begin{table}[h]
\vspace{-2ex}
    % \color{blue}
  \centering
  \caption{Rendering Throughput Comparison}\label{tab:rendering_fps_benchmarks}
  \begin{tabular}{| c | c | c | c | c | c |}
      \hline
      \textbf{Sensor Resolution} & \multicolumn{5}{c|}{\textbf{Rendering Speed (FPS)}} \\ \hline
      \textit{Num. Envs.} & \textit{128} & \textit{256} & \textit{512} & \textit{1024} &  \textit{2048} \\ \hline
      \multicolumn{6}{|c|}{\textbf{Aerial Gym Simulator}} \\ \hline
      $8\times8$ & $\mathbf{2951}$ & $\mathbf{5770}$ & $\mathbf{10546}$ & $\mathbf{21690}$ & $\mathbf{37597}$ \\ \hline
      $64\times64$ & $\mathbf{2669}$ & $\mathbf{5358}$ & $\mathbf{10097}$ & $\mathbf{16501}$ & $\mathbf{26023}$ \\ \hline
      $270\times480$ & $\mathbf{1592}$ & $\mathbf{2057}$ & $\mathbf{2408}$ & $\mathbf{2627}$ & $\mathbf{2750}$ \\ \hline
        $480\times640$ & $919$ & $1053$ & $1136$ & $\mathbf{1185}$ & $\mathbf{1205}$ \\ \hline
      \multicolumn{6}{|c|}{\textbf{Isaac Gym}} \\ \hline
      $8\times8$ & $1524$ & $1971$ & $2017$ & $2169$ & $2109$ \\ \hline
      $64\times64$ & $1460$ & $1962$ & $2014$ & $2160$ & $2067$ \\ \hline
      $270\times480$ & $1430$ & $1574$ & $1875$ & $1994$ & \xmark \\ \hline
      $480\times640$ & $\mathbf{1205}$ & $\mathbf{1269}$ & $\mathbf{1350}$ & \xmark & \xmark \\ \hline
  \end{tabular}
  \vspace{-4ex}
\end{table}

Since gym-pybullet-drones~\cite{panerati2021gym_pybullet_drones} simulates quadrotors in a single environment, the simulated cameras observe other robot meshes slowing down rendering. \addition{For OmniDrones~\cite{omnidrones}, the comparison is made against the performance using both the native cameras and the NVIDIA Warp-based ray-casting functionality provided by Isaac Lab. Notably, the ray-casting functionality can only work with a single static mesh shared across environments and does not allow modifying the mesh once initialized, whereas~\ags~allows modifications to each individual environment mesh at runtime. Therefore the benchmarking is done with a heightmap provided in the respective repository}. L2F~\cite{eschmann2024learning} simulates rigid multirotors providing a large speedup compared to other simulators, but does not consider interactions with other objects and does not offer any rendering capabilities. \addition{In contrast, our implementation not only allows for highly-parallelized simulation of arbitrary multirotor systems but also provides a powerful rendering implementation that allows maintaining and updating individual meshes per environment, offering significantly greater control over the randomization of the environments.}

We also compare the proposed rendering framework against~\isaacgym's native rendering framework for depth and segmentation images in a room-like static environment consisting of $15$ floating obstacles. The rendering operations are performed with controller-in-the-loop (to emulate practical use-cases). The rendering throughput is measured for parallel environments and tabulated in~\cref{tab:rendering_fps_benchmarks}. The~\ags~offers an order of magnitude speedup over~\isaacgym~for higher numbers of parallel environments but scales worse with resolution. Interestingly,~\isaacgym~maintains similar throughput for higher-resolution cameras \addition{across the range of parallel environments}, but requires significantly more GPU memory and crashes for a higher number of environments as indicated by \xmark~in~\cref{tab:rendering_fps_benchmarks}.

%% file: sections/evaluation/experiments.tex
We evaluate the performance of the proposed simulator for training policies for various levels of control abstraction and sensor configurations. \addition{We use the RL Games framework~\cite{rlgames2021} to train policies for high-level control tasks and simultaneously use Sample Factory~\cite{petrenko2020sample} to train policies for both low-level control and vision-based navigation tasks across different robot platforms. The robot inertias are obtained using CAD, while the response of the simulated controllers and motor models is matched with that of the real-world system. The trained policies are directly deployed on real-world robots without additional fine-tuning to evaluate their performance}. All evaluations presented are instances of open-sourced examples integrated with the~\ags.

\subsection{Position-Setpoint Tracking Task}\label{subsubsec:position_setpoint_task}

\begin{figure}[!ht]
    \centering
    \includegraphics[width=0.98\columnwidth]{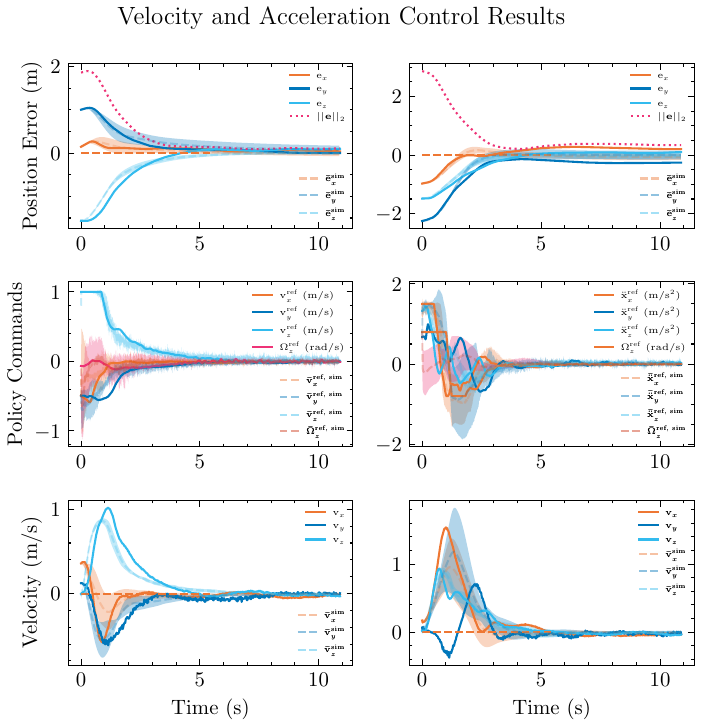}
    \vspace{-1ex}
    \caption{\addition{Real-world testing of position-setpoint tracking policies with velocity (left) and acceleration (right) commands. Solid lines show data from the real-world experiment, while the dashed lines and the shaded regions indicate mean and standard deviation across policies evaluated in simulation.}}\label{fig:sim2real_position_setpoint}
    \vspace{-1ex}
\end{figure}

We train state-based \ac{drl} policies to control the position of a quadrotor by commanding it $3$-D velocity and acceleration commands, alongside yaw rate commands. \addition{The platform similar to~\cite{kulkarni2024rl-collvae} is used, with ArduPilot firmware and onboard odometry estimation based on ROVIO~\cite{bloesch2015rovio} using an Intel RealSense Depth Camera D455.} The problem is formulated as a Markov Decision Process where the state of the robot $\mathbf{s}$ at time $t$ is defined as $\mathbf{s}_t = [\mathbf{e}_t, \addition{\mathbf{q}_t}, \mathbf{v}_t, \boldsymbol{\Omega}_t, \mathbf{a}_{t-1}]$, with $\mathbf{e}_t$ being the $3$-D position-error expressed in $\mathcal{W}$ and \addition{$\mathbf{q}_t$}  representing the orientation of the robot, $\mathbf{v}_t$ and $\boldsymbol{\Omega}_t$ are the linear and angular velocities of the robot expressed in $\mathcal{B}$. We use $x, y, z$ as subscripts denoting the values for the respective axes. $\mathbf{a}_{t-1}$ is the commanded action to the robot in the previous time-step. The action at time $t$ is defined as $\mathbf{a}_t = [\mathbf{v}_{t}^{\textrm{ref}},{\Omega}^{\textrm{ref}}_{z, t}]$ for velocity-control task and $\mathbf{a}_t = [\ddot{\mathbf{x}}^{\textrm{ref}}_{t}, {\Omega}^{\textrm{ref}}_{z, t}]$ for acceleration-control task, where $\mathbf{v}^{\textrm{ref}}_{t}$, $\ddot{\mathbf{x}}_{t}^{\textrm{ref}}$ and ${\Omega}^{\textrm{ref}}_{z, t}$ are the $3$-D velocity, $3$-D acceleration and yaw rate commands respectively, expressed in the vehicle-frame $\mathcal{V}$. ~\cref{fig:sim2real_position_setpoint} shows the real-world performance of the policies.

\subsection{Vision-based Navigation Task}\label{subsubsec:vision_based_navigation_task}

\begin{figure}[!h]
\vspace{-1ex}
    \centering
    \includegraphics[width=0.98\columnwidth]{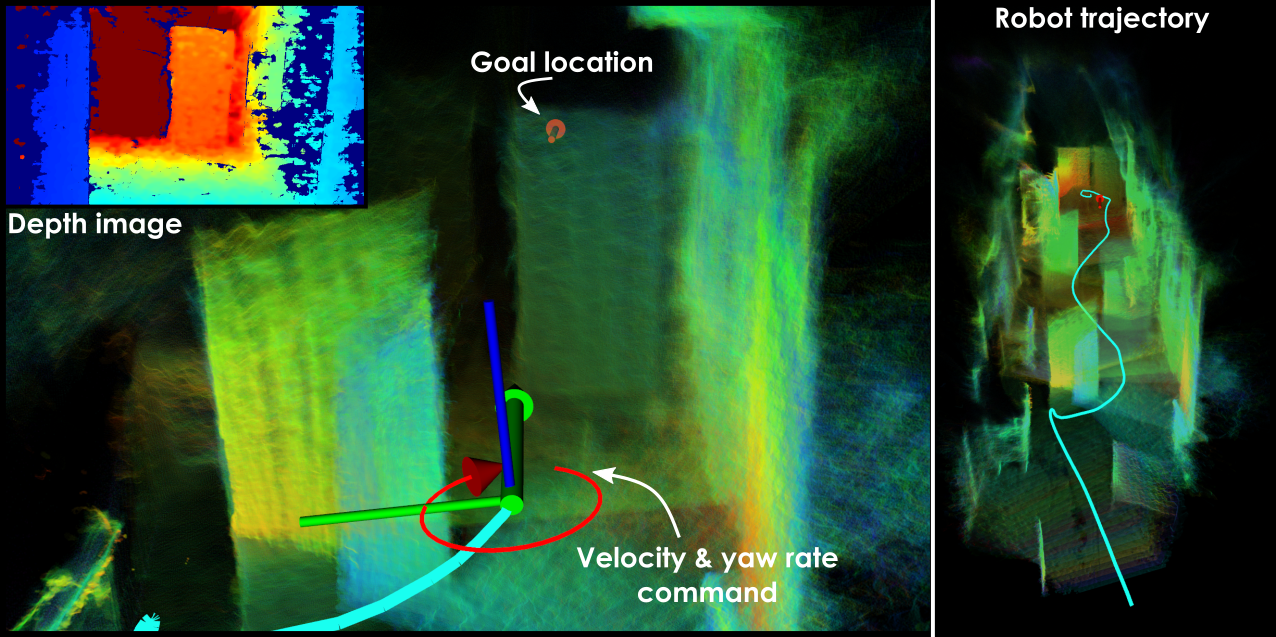}
    \caption{\addition{Real-world deployment of the} depth-image based, map-free navigation \addition{policy} in a cluttered corridor. The policy provides commands to navigate the environment safely despite sensor noise.}\label{fig:sim2real_navigation}
    \vspace{-4ex}
\end{figure}

Demonstrating the applicability of our simulator for sim2real transfer for tasks involving high-dimensional and noisy depth data, we train a policy for fast, map-free navigation of cluttered environments. We define the problem similar to the work in~\cite{kulkarni2024rl-collvae}, and train a policy with the latent space from the Deep Collision Encoder (DCE)~\cite{dce_isvc_2023}. The state is defined as $\mathbf{s}_t = [\mathbf{n}_t/||\mathbf{n}_t||_2, ||\mathbf{n}_t||_2, \Theta, \Psi, \mathbf{v}_t, \boldsymbol{\Omega}_t, h(\mathbf{a}_{t-1}), \mathbf{z}_t]$, where $\mathbf{n}_t$ represents $\mathbf{e}_t$ expressed in $\mathcal{V}$, $\Theta$ and $\Psi$ represent robot roll and pitch angles, while $h(\cdot)$ is a transformation to obtain velocity and yaw rate commands, identical to the transformation in~\cite{kulkarni2024rl-collvae}. $\mathbf{z}_t$ is a $64$-dimensional representation obtained from a $848\times480$ resolution depth image using the DCE. The policy is trained for a wall-clock time of $142~\textrm{min}$. \addition{During real-world testing using the above platform}, we constrain the vertical velocity commanded by the network to discourage the robot from going over the obstacles. Instances of policy commands along with robot path are visualized in~\cref{fig:sim2real_navigation}. With a maximum and mean speeds of $2.83~\textrm{m/s}$ and $1.54~\textrm{m/s}$ respectively, the robot navigates the cluttered corridor despite significant noise in the depth image.

\subsection{Motor Control for Position-Setpoint Tracking Task}
\begin{figure}[!ht]
    \centering
    \includegraphics[width=0.98\columnwidth]{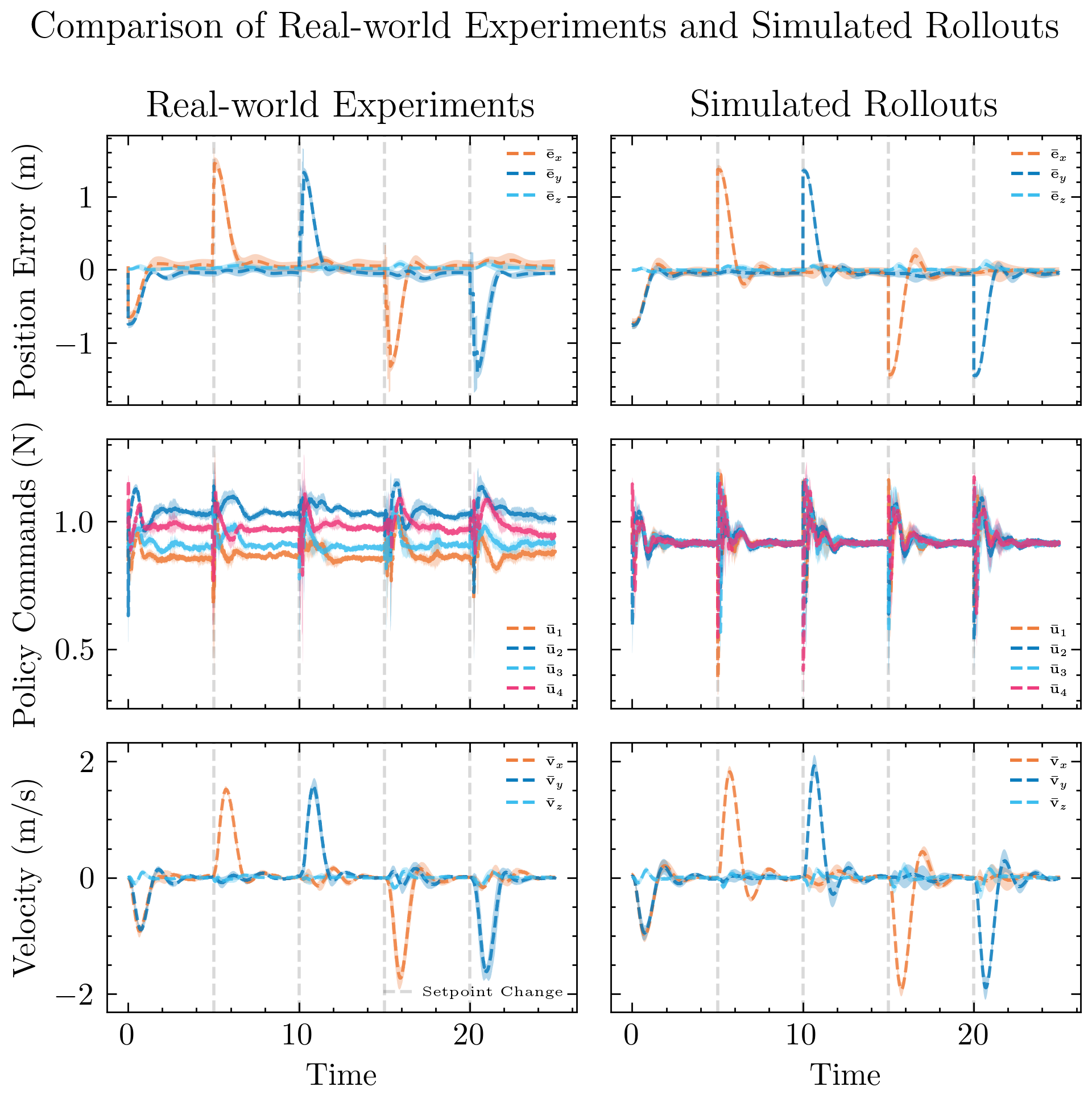}
    \vspace{-2ex}
    \caption{\addition{Mean and standard deviation (shaded) of state observations and commanded actions across motor-control policies trained with $5$ different seeds. Experiments with identical setpoint sequences are evaluated in both simulation and the real-world.}}
    \label{fig:sim2real-thrust}
    \vspace{-3ex}
\end{figure}

\addition{
We use a different planar quadrotor platform equipped with the ModalAI Voxl 2 Mini board and ModalAI Voxl ESC. A Qualisys Motion Capture system is used to estimate the pose of the robot. A fully-connected neural network consisting of $2$-layers with $32$ and $24$ neurons is trained using Sample Factory~\cite{petrenko2020sample} for $5$ different seeds. The network is deployed on the compute board in a custom PX4 module using Eigen and C++. The state is defined as $\mathbf{s}_t = [\mathbf{e}^{\mathcal{W}}_t, \mathbf{R}_6, \mathbf{R}\mathbf{v}_t, \boldsymbol{\Omega}_t]$, where $\mathbf{e}^{\mathcal{W}}_t$ denotes the position error expressed in $\mathcal{W}$, while $\mathbf{R}_6$ denotes the 6-D rotation representation based on~\cite{zhou2019continuity}. The network commands motor thrust setpoints $\mathbf{a}_t = [\mathbf{U}^{\textrm{ref}}_{t}]$. These are converted to RPM setpoints using identical thrust constants and commanded to the motors. We deploy all policies to evaluate and compare their performance. Figure~\ref{fig:sim2real-thrust} shows the mean and standard deviation of measurements across experiments.
}

\addition{
The networks are trained with a control time-step duration of $0.01~\textrm{s}$ in simulation. Inference is performed at $250~\textrm{Hz}$. The real-world performance matches the simulated rollouts with high-repeatability, with the average position tracking error across seeds $||\mathbf{e}||_2 = 0.09~\textrm{m}$. The policy commands in the simulated experiment converge to the same value for all motors, while the real-world thrust commands differ across motors. This might occur due to asymmetric mass distribution on the real-world platform leading to unmodeled torque or due to varying motor constants for different motors. Nevertheless, we achieve stable real-world tracking performance demonstrating the robustness enabled by the simulation framework.
}

%% file: sections/conclusions.tex
\section{CONCLUSIONS}\label{sec:conclusions}

\addition{
This paper presented the~\ags, an integrated simulation and rendering framework for multirotor platforms. Aerial Gym introduces functionality to support simulation and control of airframes with an arbitrary number of motors. A custom rendering framework is developed for accelerated ray-casting, exploits custom sensor models and supports parallelization across unique environment instances and offers an order of magnitude speedup compared to~\isaacgym. Extensive studies are conducted to demonstrate the robustness and sim2real transferability of state-based control and vision-based navigation policies. The simulator is open-sourced at:\\ {\url{https://github.com/ntnu-arl/aerial_gym_simulator}
}}.